\documentclass[12pt, final]{l4dc2023}

\usepackage{amsmath,amsfonts,bm}

\def\eqref#1{equation~\ref{#1}}
\def\Eqref#1{Equation~\ref{#1}}

\def\1{\bm{1}}

\def\vzero{{\bm{0}}}

\def\vd{{\bm{d}}}

\def\vp{{\bm{p}}}
\def\vq{{\bm{q}}}

\def\vu{{\bm{u}}}

\def\vx{{\bm{x}}}
\def\vy{{\bm{y}}}
\def\vz{{\bm{z}}}

\def\mA{{\bm{A}}}

\def\mC{{\bm{C}}}

\def\mG{{\bm{G}}}

\def\mJ{{\bm{J}}}
\def\mK{{\bm{K}}}
\def\mL{{\bm{L}}}

\def\mR{{\bm{R}}}

\DeclareMathAlphabet{\mathsfit}{\encodingdefault}{\sfdefault}{m}{sl}
\SetMathAlphabet{\mathsfit}{bold}{\encodingdefault}{\sfdefault}{bx}{n}

\usepackage{mathtools}

\usepackage{hyperref}
\usepackage{enumitem}

\usepackage{pgfplots}
\usepgfplotslibrary{groupplots,dateplot}
\usetikzlibrary{patterns,shapes.arrows}
\pgfplotsset{compat=newest}
\usepackage{tikz}
\usetikzlibrary{shapes, arrows.meta, positioning}

\usepgfplotslibrary{groupplots}
\usepgfplotslibrary{fillbetween}
\usetikzlibrary{arrows,decorations.pathmorphing,positioning,fit,trees,shapes,shadows,automata,calc} 
\usetikzlibrary{patterns,arrows,arrows.meta,calc,shapes,shadows,decorations.pathmorphing,decorations.pathreplacing,automata,shapes.multipart,positioning,shapes.geometric,fit,circuits,trees,shapes.gates.logic.US,fit, matrix,arrows.meta, quotes}
\usetikzlibrary{backgrounds,scopes}
\usepackage{neuralnetwork}
\usepackage{listofitems} 

\usepackage{wrapfig}

\usepackage{adjustbox}
\usepackage{multirow}
\usepackage{pbox}

\usepgfplotslibrary{external}
\pgfplotsset{compat=newest}
\newenvironment{customlegend}[1][]{%
	\begingroup
	\csname pgfplots@init@cleared@structures\endcsname
	\pgfplotsset{#1}%
}{%
	\csname pgfplots@createlegend\endcsname
	\endgroup
}%

\def\addlegendimage{\csname pgfplots@addlegendimage\endcsname}

\usepackage{times}
\title[Learning Compositional Models Using Port-Hamiltonian Neural Networks]{Compositional Learning of Dynamical System Models Using \\Port-Hamiltonian Neural Networks}

\author{%
 \Name{Cyrus Neary} \Email{cneary@utexas.edu}\\
 \Name{Ufuk Topcu} \Email{utopcu@utexas.edu}\\
 \addr The University of Texas at Austin, United States\\
}

\pgfdeclarelayer{background2}
\pgfdeclarelayer{background1}
\pgfdeclarelayer{foreground1}
\pgfdeclarelayer{foreground2}
\pgfsetlayers{background2,background1,main,foreground1,foreground2}

\begin{document}

\maketitle

\definecolor{submodel1Color}{HTML}{E05F15}
\definecolor{submodel2Color}{HTML}{07742D}
\definecolor{compositeModelColor}{HTML}{4F359B}
\definecolor{trueModelColor}{HTML}{130303}

\definecolor{lightSubmodel1Color}{HTML}{F19B6A}
\definecolor{lightSubmodel2Color}{HTML}{0cd452}
\definecolor{lightCompositeModelColor}{HTML}{a494c4}

\newcommand{\defeq}{\vcentcolon=}
\newcommand{\eqdef}{=\vcentcolon}
\newcommand{\norm}[1]{\left\lVert#1\right\rVert}

\newcommand{\state}{\vx}
\newcommand{\stateDim}{n}
\newcommand{\stateSpace}{\mathbb{R}^{\stateDim}}
\newcommand{\controlInput}{\vu}
\newcommand{\controlDim}{m}
\newcommand{\controlSpace}{\mathbb{R}^{\controlDim}}
\newcommand{\controlOutput}{\vy}
\newcommand{\interactionInput}{\vd}
\newcommand{\interactionOutput}{\vz}

\newcommand{\system}{\mathcal{S}}
\newcommand{\phs}{PHS}
\newcommand{\phsRHS}{P}

\newcommand{\generalizedPosition}{\vq}
\newcommand{\generalizedMomentum}{\vp}

\newcommand{\comp}{c}

\newcommand{\compositeState}{\state_{\comp}}
\newcommand{\compositeControl}{\controlInput_{\comp}}

\newcommand{\timeVar}{t}
\newcommand{\dt}{\Delta \timeVar}

\newcommand{\hamiltonian}{H}
\newcommand{\structure}{\mJ}
\newcommand{\dissipation}{\mR}
\newcommand{\inputControlMatrix}{\mG}
\newcommand{\inputInteractionMatrix}{\mK}

\newcommand{\params}{\theta}
\newcommand{\nnParams}{\theta}
\newcommand{\numSubmodels}{k}

\newcommand{\phnn}{PHNN}
\newcommand{\phnnRHS}{\phsRHS}
\newcommand{\phnode}{\phnn}
\newcommand{\allParams}{\Theta}

\newcommand{\data}{\mathcal{D}}
\newcommand{\trajectory}{\tau}

\newcommand{\springConstant}{k}
\newcommand{\dampingConstant}{b}
\newcommand{\mass}{m}

\newcommand{\vectorField}{F}
\newcommand{\unknownTerm}{g}
\newcommand{\collectionUnknownTerms}{G}

\newcommand{\dataset}{\mathcal{D}}
\newcommand{\datasetDim}{|\dataset|}
\newcommand{\trajectoryTimeHorizon}{|\trajectory|}
\newcommand{\rollout}{v}
\newcommand{\initState}{\state_{0}}
\newcommand{\initTime}{\timeVar_{0}}
\newcommand{\predictionTime}{T}
\newcommand{\funk}{g}
\newcommand{\numUnknownTerms}{l}

\newcommand{\nnParamsAll}{\allParams}

\newcommand{\loss}{\mathcal{L}}

\newcommand{\compositionTermParams}{\phi}

\newcommand{\diag}{\textrm{Diag}}

\begin{abstract}
Many dynamical systems---from robots interacting with their surroundings to large-scale multiphysics systems---involve a number of interacting subsystems.
Toward the objective of learning composite models of such systems from data, we present i) a framework for compositional neural networks, ii) algorithms to train these models, iii) a method to compose the learned models, iv) theoretical results that bound the error of the resulting composite models, and v) a method to learn the composition itself, when it is not known a priori.
The end result is a modular approach to learning: neural network submodels are trained on trajectory data generated by relatively simple subsystems, and the dynamics of more complex composite systems are then predicted without requiring additional data generated by the composite systems themselves.
We achieve this compositionality by representing the system of interest, as well as each of its subsystems, as a \textit{port-Hamiltonian neural network }(PHNN)---a class of neural ordinary differential equations that uses the port-Hamiltonian systems formulation as inductive bias.
We compose collections of PHNNs by using the system's physics-informed \textit{interconnection structure}, which may be known a priori, or may itself be learned from data.
We demonstrate the novel capabilities of the proposed framework through numerical examples involving interacting spring-mass-damper systems.
Models of these systems, which include nonlinear energy dissipation and control inputs, are learned independently.
Accurate compositions are learned using an amount of training data that is negligible in comparison with that required to train a new model from scratch.
Finally, we observe that the composite PHNNs enjoy properties of port-Hamiltonian systems, such as cyclo-passivity---a property that is useful for control purposes.
\end{abstract}


\section{Introduction}

Deep learning methods that use physics-based knowledge as inductive bias have recently shown promise in learning dynamical system models that respect physical laws and that generalize beyond the training dataset~\citep{djeumou2022neural, menda2019structured, gupta2020structured, cranmer2020lagrangian, greydanus2019hamiltonian, finzi2020simplifying, zhong2021benchmarking}.
These methods, which often use neural networks to parametrize select terms in differential operators, are able to learn complex relationships from data while also yielding models that are compact and interpretable.

However, there remain barriers to the deployment of such algorithms in engineering applications.
Many systems, from robots interacting with their surroundings to large-scale multiphysics systems, involve large numbers of interacting components.
These interactions between subsystems can increase the complexity of the overall system's dynamics, rendering monolithic approaches to learning---which capture the entire system using a single model learned from data---challenging.

\begin{figure}
    \centering
    \input{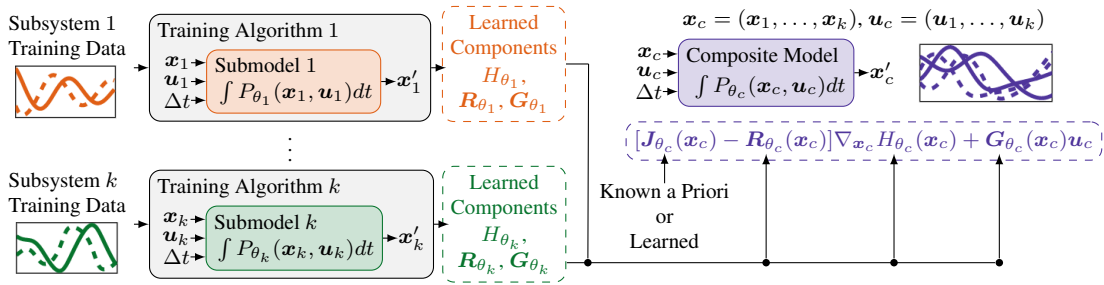}
    \caption{
        An illustration of the compositional learning framework. 
        We train separate port-Hamiltonian neural networks (PHNNs) on data generated by individual subsystems, presented in \S \ref{sec:phnns}.
        We then compose these submodels to construct another PHNN that models the composite system, presented in \S \ref{sec:composing_phnn}.
    }
    \vspace*{-5mm}
    \label{fig:intro_flowchart}
\end{figure}

We present a framework and algorithms for learning and composing neural network models of dynamical systems.
The framework models individual subsystems independently, and uses physics-informed interfaces between these submodels to capture their interactions.
This compositional approach to learning provides a number benefits and novel capabilities that would not otherwise be possible.
Firstly, it simplifies the learning problems to be solved.
Submodels are trained on trajectory data generated by relatively simple subsystems.
The dynamics of more complex composite systems are then predicted without requiring additional training.
Secondly, it provides a modular framework for data-driven modeling. Previously learned component models can be composed in new ways to simulate different composite systems.
Finally, it provides a natural way to compose data-driven models with models derived from first principles.

We achieve this compositionality by representing the system of interest, as well as each of its subsystems, as a \textit{port-Hamiltonian neural network }(PHNN)---a class of deep learning models that use the port-Hamiltonian systems formalism \citep{duindam2009modeling, van2014port} to inform the model's structure.
More specifically, PHNNs parametrize each subsystem's Hamiltonian function, as well as how it dissipates energy, interacts with other subsystems, and how it responds to control inputs.
We enforce known properties of the dissipation and interaction terms through the model's construction.
The PHNN's output is obtained by numerically integrating a differential equation involving all of these terms to predict the system's state at a future time.

Using the physics-informed structure provided by the PHNN, we present a method to compose collections of PHNNs in order to obtain models of the corresponding composite systems, and we provide upper bounds on this composite model's prediction errors.
Figure \ref{fig:intro_flowchart} illustrates the approach.
The composite system's Hamiltonian, dissipation term, and control input term are all obtained by combining the corresponding terms from the learned submodels.
Interactions between the subsystems are captured by the \textit{interconnection structure} of the composite system, which may be known a priori, or may itself be learned from data.
In many cases this interconnection structure is given by a constant linear operator; we accordingly present a method to learn it via linear regression. 
In the general case, it may instead be parameterized using a neural network.

We demonstrate the novel capabilities of the proposed framework through numerical examples involving interacting spring-mass-damper systems.
Models of these systems, which include nonlinear energy dissipation and control inputs, are learned independently.
The dynamics of the composite system are accurately predicted without additional training. 
If the system's interconnection structure is unknown, we demonstrate that an accurate composition may be learned using an amount of training data that is negligible in comparison with that required to train a new model from scratch.
Finally, we empirically observe that the proposed compositions of PHNNs exhibit the property of passivity---a property of port-Hamiltonian systems that is useful for control purposes.

\section{Related Work}
\label{sec:related_work}

The port-Hamiltonian formulation of dynamical systems provides a rich mathematical framework that enables compositional modeling \citep{duindam2009modeling, van2014port}.
This framework can be applied quite generally and has been used to model fluid-structure interactions \citep{cardoso2015modeling}, aerial vehicles in contact scenarios \citep{rashad2021energy}, and coupled gas and electricity distribution networks \citep{strehle2018towards}.
Furthermore, a wealth of existing methods and analysis for the nonlinear control of port-Hamiltonian systems already exists \citep{van2020port}.
However, deriving a precise system model in port-Hamiltonian form can be challenging in many practical applications  \citep{nageshrao2015adaptive, cherifi2019numerical}.
Furthermore, methods for data-driven identification and control of port-Hamiltonian systems have not yet been extensively explored \citep{nageshrao2015adaptive, cherifi2022non}.

The inclusion of physics-based knowledge into neural network models of dynamical systems has, however, been studied extensively over the past several years.
In particular, neural ordinary differential equations (NODEs) \citep{chen2018neural} provide a natural approach to incorporate physics-based knowledge as inductive bias in deep learning \citep{zhong2021benchmarking, rackauckas2020universal}. 
By using neural networks to parametrize differential equations, as opposed to directly fitting the available trajectory data, NODEs allow the user to harness an existing wealth of knowledge from applied mathematics, physics, and engineering \citep{djeumou2022neural, cranmer2020lagrangian, lutter2019deep, gupta2020structured, roehrl2020modeling, zhong2021differentiable, shi2019neural}.

Of particular relevance to our work, \textit{Hamiltonian neural networks} use the Hamiltonian formulation of dynamics to inform the structure of a neural ODE \citep{greydanus2019hamiltonian, matsubara2019deep, toth2019hamiltonian, finzi2020simplifying}.
However, Hamiltonian-based models necessarily represent closed systems.
By contrast, we study systems involving energy exchange, energy dissipation, and control inputs.
Meanwhile, \cite{xu2021neural, furieri2022distributed, plaza2022total} use neural networks to parametrize controllers for port-Hamiltonian systems.
However, these works do not learn dynamics models---they assume the dynamics to be known and focus on control.

More closely related to our work, several recent papers also study neural ODEs that have a port-Hamiltonian structure \citep{zhong2020dissipative, desai2021port, eidnes2022port, duong2021hamiltonian}.
However, none of these works study how physics-based knowledge may be used to compose deep learning models.
By contrast, the primary focus of this work is to develop a framework, theoretical results, and methods that enable such compositional learning algorithms.
\section{Port-Hamiltonian Systems}
\label{sec:phs}

Port-Hamiltonian (PH) systems provide a versatile framework that enables the modeling and analysis of complex networks of interacting subsystems.
Conceptually, PH systems are represented by their Hamiltonian functions, by their energy dissipation terms, and by a mathematical description of the power-conserving interactions of their subsystems, called a \textit{Dirac structure}.
We refer to \cite{van2014port, duindam2009modeling} for further details.

In this work, we consider lumped parameter PH systems expressed in \textit{explicit state-input-output} form; the system's state may be represented as a finite-dimensional vector, and its dynamics are given by \eqref{eq:state_input_output_ph_form_state}.
PH systems may be expressed in this form whenever there are no algebraic constraints on the system's state variables \citep{donaire2009derivation, dai2019model}.
\begin{equation}
    \dot{\state} = \left[ \structure(\state) - \dissipation(\state) \right] \nabla_{\state} \hamiltonian(\state) + \inputControlMatrix(\state) \controlInput, \;\;\; \controlOutput = \inputControlMatrix(\state)^{T} \nabla_{\state} \hamiltonian(\state) \label{eq:state_input_output_ph_form_state} \\
\end{equation}
Here, \(\state \in \stateSpace\) is the \(\stateDim\)-dimensional vector representing the system's state, \(\hamiltonian(\state)\) is the system's Hamiltonian function, 
\(\structure(\state) \in \mathbb{R}^{\stateDim \times \stateDim}\) is the skew symmetric interconnection matrix (i.e. \(\structure(\state) = - \structure(\state)^{T}\) for every \(\state \in \stateSpace\)), 
\(\dissipation(\state) \in \mathbb{R}^{\stateDim \times \stateDim}\) is the symmetric positive semi-definite dissipation matrix (i.e. \(\dissipation(\state) = \dissipation(\state)^{T}\) and \(\Tilde{\state}^T \dissipation(\state) \Tilde{\state} \geq 0\) for every \(\state, \Tilde{\state} \in \stateSpace\)),
\(\inputControlMatrix(\state) \in \mathbb{R}^{\stateDim \times \controlDim}\) is the control input matrix,
\(\controlInput \in \controlSpace\) is the \(\controlDim\)-dimensional control input vector, and \(\controlOutput \in \controlSpace\) is the corresponding output vector.
Intuitively, while \(\hamiltonian(\state)\) describes the system's energy in terms of its state \(\state\), \(\structure(\state)\) encodes the energy-conserving interactions between the various components of the system and \(\dissipation(\state)\) encodes how these components dissipate energy.

\subsection{An Illustrative Running Example: The Coupled Mass-Spring-Damper}
\label{sec:illustrative_example}

\begin{figure}
    \centering
    \vspace{-3mm}
    \begin{tikzpicture}[black!75,thick]
 
\newlength{\wallOffset}
\newlength{\wallWidth}
\newlength{\wallHeight}
\newlength{\wallCompositeHeight}
\newlength{\springOneWidth}
\newlength{\springTwoWidth}
\newlength{\massHeight}
\newlength{\massWidth}
\newlength{\damperHeight}
\newlength{\damperWidth}
\newlength{\groundWidth}
\newlength{\groundHeight}
\newlength{\springDampSep}
\newlength{\forceVecLength}

\setlength{\wallOffset}{-3cm}
\setlength{\wallWidth}{0.2cm}
\setlength{\wallHeight}{0.75cm}
\setlength{\wallCompositeHeight}{1.5cm}
\setlength{\springOneWidth}{2cm}
\setlength{\springTwoWidth}{1.6cm}
\setlength{\massHeight}{0.75cm}
\setlength{\massWidth}{1cm}
\setlength{\damperHeight}{0.1cm}
\setlength{\damperWidth}{0.2cm}
\setlength{\groundWidth}{3.5cm}
\setlength{\groundHeight}{0.2cm}
\setlength{\springDampSep}{0.3cm}
\setlength{\forceVecLength}{0.5cm}

\tikzstyle{spring}=[
    decorate,
    decoration={
        coil,
        aspect=0.3, 
        segment length=1.2mm, 
        amplitude=1mm, 
        pre length=3mm,
        post length=3mm,
    }
]

\tikzstyle{damper}=[
    decorate,
    thick,
    decoration={
        markings,  
        mark connection node=dmp,
        mark=at position 0.5 with {
            \node (dmp) [
                thick,
                inner sep=0pt,
                transform shape,
                rotate=-90,
                minimum width=\damperWidth,
                minimum height=\damperHeight,
                draw=none
            ] {};a
            \draw [thick] ($(dmp.north east)+(2pt,0)$) -- (dmp.south east) -- (dmp.south west) -- ($(dmp.north west)+(2pt,0)$);
            \draw [thick] ($(dmp.north)+(0,-3pt)$) -- ($(dmp.north)+(0,3pt)$);
        }
    }, 
]

\node [
    pattern = north east lines,
    minimum width=\wallWidth,
    minimum height=\wallHeight,
] (wall1) {};
\draw[thick] ($(wall1.east) + (0, -\wallHeight / 2)$) -- ($(wall1.east) + (0,\wallHeight / 2)$);

\node[
    pattern = north east lines,
    minimum width=\groundWidth,
    minimum height=\groundHeight,
    below right = \wallHeight/2 and 0.0cm of wall1.west,
] (ground1) {};
\draw[thick] ($(wall1.east) + (0, -\wallHeight / 2)$) -- ($(ground1.east) + (0.0, \groundHeight / 2)$);

 
\node[draw,
    color=submodel1Color,
    fill=submodel1Color,
    minimum width=1cm,
    minimum height=0.75cm,
    anchor=west,
    right = \springOneWidth of wall1,
    text=white] (M1) {$m_{1}$};
 
\draw[spring, color=submodel1Color] ($(wall1.east) + (0.0, \springDampSep / 2)$) -- ($(M1.west) + (0.0, \springDampSep / 2)$); 

\draw[damper, color=submodel1Color] ($(wall1.east) + (0.0, -\springDampSep / 2)$) -- ($(M1.west) + (0.0, -\springDampSep / 2)$); 
 
\node[above=0.65cm of ground1.north, color=submodel1Color] (subsystem1Label) {Subsystem 1};
 
 
\node [
    pattern = north east lines,
    minimum width=\wallWidth,
    minimum height=\wallHeight,
    below = 0.7cm of wall1.south,
] (wall2) {};
\draw[thick] ($(wall2.east) + (0, -\wallHeight / 2)$) -- ($(wall2.east) + (0,\wallHeight / 2)$);

\node[
    pattern = north east lines,
    minimum width=\groundWidth,
    minimum height=\groundHeight,
    below right = \wallHeight/2 and 0.0cm of wall2.west,
] (ground2) {};
\draw[thick] ($(wall2.east) + (0, -\wallHeight / 2)$) -- ($(ground2.east) + (0.0, \groundHeight / 2)$);
 
\node[draw,
    color=submodel2Color,
    fill=submodel2Color,
    minimum width=1cm,
    minimum height=0.75cm,
    anchor=west,
    right = \springTwoWidth of wall2,
    text=white,] (M2) {$m_{2}$};
 
\draw[spring, color=submodel2Color] ($(wall2.east) + (0.0, \springDampSep / 2)$) -- ($(M2.west) + (0.0, \springDampSep / 2)$); 

\draw[damper, color=submodel2Color] ($(wall2.east) + (0.0, -\springDampSep / 2)$) -- ($(M2.west) + (0.0, -\springDampSep / 2)$); 
 
\draw[
    very thick,
    submodel2Color,
    -latex
] (M2.east) -- ++(\forceVecLength, 0.0) node[above]{\(\controlInput_{2}\)};
 
\node[above=0.65cm of ground2.north, color=submodel2Color] (subsystem2Label) {Subsystem 2};


\node [
    pattern = north east lines,
    minimum width=\wallWidth,
    minimum height=\wallCompositeHeight,
    above right = -0.39cm and 4cm of wall2.east,
] (wall3) {};
\draw[thick] ($(wall3.east) + (0, -\wallCompositeHeight / 2)$) -- ($(wall3.east) + (0,\wallCompositeHeight / 2)$);

\node[
    pattern = north east lines,
    minimum width=2*\groundWidth,
    minimum height=\groundHeight,
    below right = \wallCompositeHeight/2 and 0.0cm of wall3.west,
] (ground3) {};
\draw[thick] ($(wall3.east) + (0, -\wallCompositeHeight / 2)$) -- ($(ground3.east) + (0.0, \groundHeight / 2)$);
 
\node[draw,
    color=submodel1Color,
    fill=submodel1Color,
    minimum width=1cm,
    minimum height=0.75cm,
    anchor=west,
    below right = \wallCompositeHeight/2 - \massHeight and \springOneWidth of wall3.east,
    text=white] (M1comp) {$m_{1}$};
 
\node[draw,
    color=submodel2Color,
    fill=submodel2Color,
    minimum width=1cm,
    minimum height=0.75cm,
    anchor=west,
    right = \springTwoWidth of M1comp.east,
    text=white,] (M2comp) {$m_{2}$};
 
\draw[spring, color=submodel1Color] ($(M1comp.west) + (-\springOneWidth, \springDampSep / 2)$) -- ($(M1comp.west) + (0.0, \springDampSep / 2)$);

\draw[damper, color=submodel1Color] ($(M1comp.west) + (-\springOneWidth, -\wallCompositeHeight/2 + \massHeight -\springDampSep / 2)$) -- ($(M1comp.west) + (0.0, -\springDampSep / 2)$);
 
\draw[spring, color=submodel2Color] ($(M1comp.east) + (0.0, \springDampSep / 2)$) -- ($(M2comp.west) + (0.0, \springDampSep / 2)$); 

\draw[damper, color=submodel2Color] ($(M1comp.east) + (0.0, -\springDampSep / 2)$) -- ($(M2comp.west) + (0.0, -\springDampSep / 2)$); 
 
\draw[
    very thick,
    submodel2Color,
    -latex
] (M2comp.east) -- ++(\forceVecLength, 0.0) node[above]{\(\controlInput_{2}\)};
 
\node[above=1.4cm of ground3.north, color=compositeModelColor] (compSystemLabel) {Composite System};

\node[above = 2.0cm of ground3.north, align=center] (stateVariableDescription) {\footnotesize \color{submodel1Color}\(\state_{1} = (\Delta q_{1}, p_{1})\), \color{submodel2Color} \(\state_{2} = (\Delta q_{2}, p_{2})\), \color{compositeModelColor} \(\state_{\parallel} = (\state_{1}, \state_{2})\)};
\color{black}
 
\draw[gray,dashed] (M1comp.north) -- ++(0.0cm, 0.5cm);
\draw[
    gray,
    latex-latex] ($(wall3.east) + (0.0cm, 0.25cm)$) -- ($(M1comp.north) + (0.0, 0.25cm)$)
    node[midway,above]{\small $\Delta q_{1}$};

\draw[gray,dashed] (M2comp.north) -- ++(0.0cm, 0.5cm);
\draw[
    gray,
    latex-latex] ($(M1comp.north) + (0.0cm, 0.25cm)$) -- ($(M2comp.north) + (0.0, 0.25cm)$)
    node[midway,above]{\small $\Delta q_{2}$};
 
 
 
 
 
\end{tikzpicture}
    \vspace{-3mm}
    \caption{
        An illustrative coupled spring-mass-damper example.
        Two subsystems (left), with different spring and damping constants, are connected to obtain the composite system (right).
    }
    \vspace{-3mm}
    \label{fig:spring_mass_example}
\end{figure}
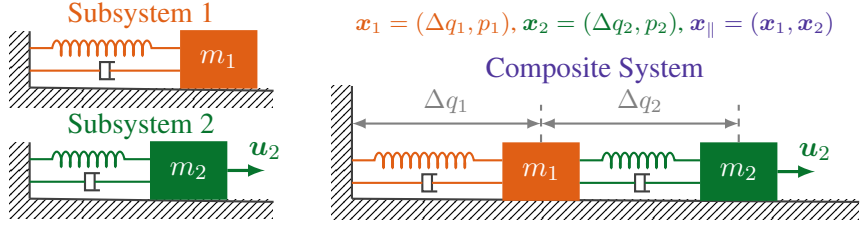

Let \(\phs_{1}\) and \(\phs_{2}\) represent the two subsystems on the left of Figure \ref{fig:spring_mass_example}, and let \(\phs_{\comp}\) represent the system resulting from their composition, illustrated on the right.
Similarly, let \(\state_{1}\), \(\state_{2}\), \(\controlInput_{1}\), and \(\controlInput_{2}\) denote the subsystem states and control inputs, and let \(\state_{\comp} \defeq (\state_{1}, \state_{2})\), \(\controlInput_{\comp} \defeq (\controlInput_{1}, \controlInput_2)\) denote the state and control input of the composite system.

We assume that the damping forces are nonlinear \(F_{damp} = \dampingConstant \dot{\generalizedPosition}^3\), similar to the example presented by \cite{lopes2015explicit}.
We define the subsystem states as \(\state_{i} = (\Delta \generalizedPosition_{i}, \generalizedMomentum_{i})\) for \(i=1,2\), where \(\Delta \generalizedPosition_{i}\) is the elongation of the spring and \(\generalizedMomentum_{i}\) is the momentum of the mass.
The dynamics of the subsystems may then be represented in the form of \eqref{eq:state_input_output_ph_form_state} with \(\hamiltonian_{i}(\state_{i}) = \frac{\generalizedMomentum_{i}^{2}}{2 \mass_{i}} + \frac{\springConstant \Delta \generalizedPosition_{i}^{2}}{2}\) and with
\begin{equation}
    \structure_i = \begin{bmatrix}0 & 1 \\ -1 & 0\end{bmatrix}, \;\;
    \dissipation_i(\state_i) = \begin{bmatrix}0 & 0 \\ 0 & \dampingConstant_{i} \frac{\generalizedMomentum_{i}^2}{\mass_{i}^{2}} \end{bmatrix}, \;\;
    \inputControlMatrix_{i} = \begin{bmatrix} 0 \\ 1 \end{bmatrix}.
    \label{eq:subsystem_spring_mass_equations}
\end{equation}
The port-Hamiltonian system \(\phs_{\comp}\) representing the system on the right of Figure \ref{fig:spring_mass_example} may be obtained by \textit{composing} \(\phs_{1}\) and \(\phs_{2}\), the port-Hamiltonian reprepresentations of the subsystems.
That is, the dynamics of \(\phs_{\comp}\) may also be written in the form of \eqref{eq:state_input_output_ph_form_state},
where the composite Hamiltonian is given by \(\hamiltonian_{\comp}(\state_{\comp}) = \hamiltonian_{1}(\state_1) + \hamiltonian_{2}(\state_{2})\), the dissipation \(\dissipation_{\comp}(\state_{\comp})\) and control input \(\inputControlMatrix_{\comp}\) matrices are obtained by stacking the matrices \(\dissipation_{i}(\state_{i})\) and \(\inputControlMatrix_{i}\) diagonally, and the composite interconnection term \(\structure_{\comp}\) is obtained by stacking \(\structure_{i}\) diagonally and by including an additional pair of off-diagonal interaction terms.
These additional entries in \(\structure_{\comp}\) encode the coupling between the subsystems; spring 2 exerts a force of \(\frac{\partial H_{2}}{\partial \Delta \generalizedPosition_{2}}\) on mass 1, and mass 1's contribution to the rate of change in the elongation of spring 2 is given by \(-\frac{\partial H_{1}}{\partial \generalizedMomentum_{1}}\).
The composite system's dynamics are given explicitly in Appendix \ref{sec:appendix_experimental_details}.
\section{Port-Hamiltonian Neural Networks}
\label{sec:phnns}

Our objective is to train port-Hamiltonian neural networks (PHNN) to predict the dynamics of relatively simple subsystems (e.g., the individual spring-mass-damper systems from the left of Figure \ref{fig:spring_mass_example}) and then to compose these learned models in order to simulate more complex systems (e.g., the coupled spring-mass-damper system on the right of Figure \ref{fig:spring_mass_example}).
In this section we present how to construct and train PHNN submodels.
We present a method to compose the learned models in \S \ref{sec:composing_phnn}.

\subsection{Constructing Port-Hamiltonian Neural Networks}

A PHNN parametrizes the unknown terms in \eqref{eq:state_input_output_ph_form_state}, and solves the resulting differential equation in order to predict the system's future states.
Let \(\phnnRHS(\state, \controlInput)\) denote the right-hand side of the state equation in  \ref{eq:state_input_output_ph_form_state}.
We use \(\hamiltonian_{\nnParams}(\state)\), \(\dissipation_{\nnParams}(\state)\), and \(\inputControlMatrix_{\nnParams}(\state)\) to denote the parametrizations of the potentially unknown terms in \(\phnnRHS(\state,\controlInput)\).
We use \(\phnnRHS_{\allParams}(\state, \controlInput)\) to denote the expression that results when each of the unknown terms is replaced with its parametrization, where \(\allParams\) denotes the collection of all the individual parameter vectors.

\begin{figure}
    \centering
    \vspace{-3mm}
    \input{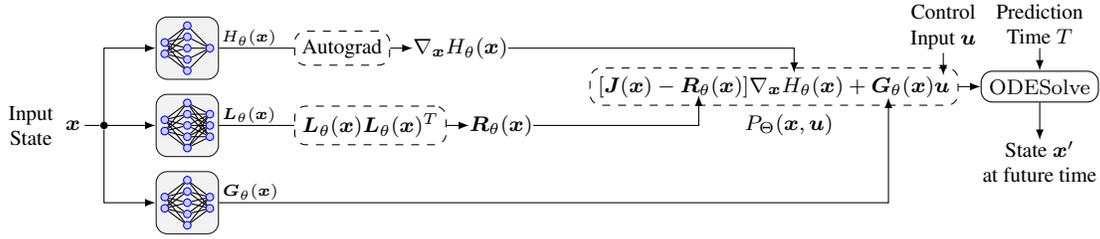}
    \vspace{-3mm}
    \caption{
        An illustration of the construction and evaluation of a port-Hamiltonian Neural Network.
    }
    \vspace{-3mm}
    \label{fig:phnode_illustration}
\end{figure}

Figure \ref{fig:phnode_illustration} illustrates the process of constructing and evaluating a PHNN.
The Hamiltonian \(\hamiltonian_{\nnParams}(\state)\) is parametrized as a multi-layer perceptron (MLP) and its gradient \(\nabla_{\state}\hamiltonian_{\nnParams}(\state)\) with respect to \(\state\) is computed using automatic differentiation.
The entries in the input matrix \(\inputControlMatrix_{\nnParams}(\state) \in \mathbb{R}^{\stateDim \times \controlDim}\) are either parametrized directly (if \(\inputControlMatrix\) is known to be a constant matrix), or as the output of an MLP (if \(\inputControlMatrix(\state)\) varies with \(\state\)).
The dissipation term \(\dissipation_{\nnParams}(\state)\) is similarly either parametrized as a constant matrix or as the output of an MLP.
However, we additionally enforce the positive semi-definiteness of \(\dissipation_{\nnParams}(\state)\) by parametrizing its Cholesky decomposition, instead of parametrizing its entries directly.
That is, similarly to as proposed by \cite{zhong2020dissipative}, we define \(\dissipation_{\nnParams}(\state) \defeq \mL_{\nnParams}(\state) \mL_{\nnParams}(\state)^{T}\) for some parametrized lower-triangular matrix \(\mL_{\nnParams}(\state) \in \mathbb{R}^{\stateDim \times \stateDim}\) with non-negative diagonal entries.

Note that in this section, similar to all existing works involving port-Hamiltonian neural networks \citep{greydanus2019hamiltonian, zhong2020dissipative, desai2021port, eidnes2022port}, we assume that the interaction term \(\structure(\state)\) is known a priori.
It is possible to learn a skew-symmetric parametrization of \(\structure(\state)\) along with all of the other terms in the PHNN.
However, doing so necessitates additional considerations that are beyond the scope of the current work, when composing the resulting models.

\paragraph{PHNNs enforce the cyclo-passivity property by construction.}

The \textit{cyclo-passivity} property enjoyed by port-Hamiltonian systems ensures that \(d\hamiltonian / d \timeVar \leq \controlOutput^{T} \controlInput\)---the rate of change of the system's energy cannot exceed the externally-provided power \citep{van2000l2}.
Regardless of the output of the parametrized Hamiltonian \(\hamiltonian_{\nnParams}(\state)\), this property is guaranteed by the skew-symmetry of \(\structure(\state)\) together with the positive semi-definiteness of the dissipation term \(\dissipation_{\nnParams}(\state)\), which we enforce through the PHNN's construction.
When dissipation and control inputs are both present, 
\(d\hamiltonian/dt = h^{T} \structure(\state) h - h^{T} \dissipation_{\nnParams}(\state) h + h^{T} \inputControlMatrix_{\nnParams}(\state) \controlInput \leq \controlOutput^{T} \controlInput\).
Here, we use \(h\) to denote \(\nabla_{\state} \hamiltonian_{\nnParams}(\state)\).

\subsection{Evaluating Port-Hamiltonian Neural Networks}
The input to a PHNN is a tuple \((\state, \controlInput, \timeVar, \predictionTime)\) consisting of the current state \(\state\), a control input \(\controlInput\), the current time \(\timeVar\), and a prediction time \(\predictionTime\) with \(\timeVar < \predictionTime\).
The output \(\hat{\state}_{\predictionTime} \defeq \phnode_{\allParams}(\state, \controlInput, \timeVar, \predictionTime)\) of the PHNN parametrized by \(\allParams\) is then given by  \(\mathrm{ODESolve}(\phnnRHS_{\allParams}(\cdot, \controlInput), \state, \timeVar, \predictionTime) \approx \state + \int_{\timeVar}^{\predictionTime} \phnnRHS_{\allParams}(\state_{s}, \controlInput)ds\), where we use the subscript notation \(\state_{\timeVar}\) to denote \(\state(\timeVar)\).
Here, \(\mathrm{ODESolve}(\phnnRHS_{\allParams}(\cdot, \controlInput), \state, \timeVar, \predictionTime)\) is a numerical solution to the ordinary differential equation specified by \(\phnnRHS_{\allParams}(\state, \controlInput)\) over the window of time \([\timeVar, \predictionTime]\).
We note that the particular algorithm used to evaluate \(\mathrm{ODESolve}(\cdot)\) influences the model's accuracy and the computational cost of forward evaluations of the model \citep{djeumou2022taylor}.
In this work, we use fixed-timestep RK4 to evaluate \(\textrm{ODESolve}(\cdot)\) in all experiments.

\subsection{Training Port-Hamiltonian Neural Networks}
We assume that a finite dataset $\dataset$ of system trajectories---time-series data of states and control inputs---is available in lieu of the system's model.
That is, we are given a set $\dataset = \{ \trajectory_1, \hdots, \trajectory_{\datasetDim}\}$ of trajectories \(\trajectory = \{(\state_0, \controlInput_0, \timeVar_{0}), \ldots, (\state_{\trajectoryTimeHorizon}, \controlInput_{\trajectoryTimeHorizon}, \timeVar_{\trajectoryTimeHorizon})\}\), where \(\state_i\) is the state at time \(t_i\), \(\controlInput_i\) is the control input applied from time \(\timeVar_{i}\) until \(\timeVar_{i+1}\), and $t_0 < t_1 < \ldots < t_{\trajectoryTimeHorizon}$ is an increasing sequence of points in time.
Note that we are using \(\phnnRHS_{\allParams}(\cdot)\) to parametrize the time derivative \(\dot{\state}\) of the system's state, which is not explicitly included in the dataset.

Given a dataset \(\dataset\), we construct the loss function \(\loss(\nnParamsAll, \dataset)\) of the PHNN to capture the error in the model's predictions of the future states, for any given norm, as
\begin{equation}
    \label{eq:loss_function}
    \loss(\allParams, \dataset) =  \sum_{\trajectory_l \in \dataset} \sum_{(\state_i, \controlInput_i, \timeVar_{i}) \in \trajectory_l} \| \phnode_{\allParams}(\state_{i}, \controlInput_{i}, \timeVar_{i}, \timeVar_{i+1}) - \state_{i+1}\|^{2}.
\end{equation} 
Finally, we search for local minima of \(\loss(\allParams, \dataset)\) using gradient-based techniques, where \(\nabla_{\allParams} \loss(\allParams, \dataset)\) may be computed using either direct automatic differentiation through \(\mathrm{ODESolve}(\cdot)\), or using the adjoint sensitivity method~\citep{pontryagin1987mathematical, chen2018neural}. 

\newcommand{\phnnErr}{\varepsilon}
\newcommand{\hamiltonianErr}{\eta}
\newcommand{\indexSet}{\mathcal{I}}
\newcommand{\compositionMatrix}{\mC}
\newcommand{\compositionMatrixNormBound}{\sigma}
\newcommand{\compositionMatrixErr}{\gamma}
\newcommand{\domain}{\Omega}

\section{Composing Port-Hamiltonian Neural Networks}
\label{sec:composing_phnn}

While in \S \ref{sec:phnns} we introduced PHNNs and methods to train them, in this section we present a method to compose previously learned PHNNs in order to predict the dynamics of larger composite systems.
In the context of the example from \S \ref{sec:illustrative_example}, we use the methods from \S \ref{sec:phnns} to learn PHNN models of the individual spring-mass-dampers, and we use the methods in this section to compose these models in order to predict the dynamics of the coupled system.

\subsection{Composing Models Using a Known Interconnection Structure}
\label{sec:composition_known_interconnection}

We construct a model of the composite system \(\phnode^{\compositionMatrix}_{\comp, \allParams_{\comp}}(\cdot)\) by combining the terms of the learned submodels \(\phnode_{i,\allParams}(\cdot)\), for \(i = 1,\ldots, \numSubmodels\), as illustrated in Figure \ref{fig:intro_flowchart}. 
Let \(\state_{i}\in \mathbb{R}^{\stateDim_{i}}\) and \(\controlInput_{i} \in \mathbb{R}^{\controlDim_{i}}\) represent the state and control input vectors of subsystem \(i\).
We define the state and control input vectors of the composite system to be \(\state_{\comp} \defeq \begin{bmatrix}\state_{1}^{T}, \ldots, \state_{\numSubmodels}^{T}\end{bmatrix}^{T} \in \mathbb{R}^{\stateDim_{\comp}}\) and \(\controlInput_{\comp} \defeq \begin{bmatrix}\controlInput_{1}^{T}, \ldots, \controlInput_{\numSubmodels}^{T}\end{bmatrix}^{T} \in \mathbb{R}^{\controlDim_{\comp}}\), where \(\stateDim_{\comp} = \stateDim_{1} + \ldots + \stateDim_{\numSubmodels}\) and \(\controlDim_{\comp} = \controlDim_{1} + \ldots + \controlDim_{\numSubmodels}\).

The Hamiltonian of the composite system is defined as the sum of the subsystem Hamiltonians, \(\hamiltonian_{\comp, \nnParams}(\state_{\comp}) \defeq \hamiltonian_{1, \nnParams}(\state_{1}) + \ldots +  \hamiltonian_{\numSubmodels, \nnParams}(\state_{\numSubmodels})\).
Note that we assume that each subsystem's Hamiltonian function \(\hamiltonian_{i}(\state_{i})\) depends only on the corresponding state vector \(\state_{i}\).
The dissipation and control input terms are defined as \(\dissipation_{\comp, \nnParams}(\state_{\comp}) \defeq \diag(\dissipation_{1,\nnParams}(\state_{1}), \ldots, \dissipation_{\numSubmodels, \nnParams}(\state_{\numSubmodels}))\) and \(\inputControlMatrix_{\comp,\nnParams}(\state_{\comp}) \defeq \diag(\inputControlMatrix_{1,\nnParams}(\state_{1}), \ldots, \inputControlMatrix_{\numSubmodels, \nnParams}(\state_{\numSubmodels}))\), respectively.
Here, we use \(\diag(\mA_{1}, \ldots, \mA_{\numSubmodels})\) to represent the matrix that results from using the submatrices \(\mA_{1}, \ldots, \mA_{\numSubmodels}\) to define the blocks of entries along the matrix diagonal, similarly to as in the example from \S \ref{sec:illustrative_example}.
Finally, the interconnection term of the composition is given by \(\structure_{\comp}(\state_{\comp}) \defeq \diag(\structure_{1}(\state_{1}), \ldots, \structure_{\numSubmodels}(\state_{\numSubmodels})) + \compositionMatrix(\state_{\comp})\),
where \(\compositionMatrix(\state_{\comp}) \in \mathbb{R}^{\stateDim_{\comp} \times \stateDim_{\comp}}\) is a skew symmetric \textit{composition} matrix encoding energy-conserving interactions between the state variables of the various subsystems.
Furthermore, we define the blocks of entries along the diagonal of \(\compositionMatrix(\state_{\comp})\) to be zero---the internal subsystem interactions defined by \(\structure_{i}(\state_{i})\) should not be altered by \(\compositionMatrix(\state_{\comp})\).
We note that the interconnection and dissipation terms of the composite model retain their properties of skew-symmetry and positive semidefiniteness, respectively.

\subsection{Learning Compositions of Port-Hamiltonian Neural Network Submodels}
\label{sec:composition_unknown_interconnection}

If the composition term \(\compositionMatrix(\state_{\comp})\) is unknown, we propose to learn it using additional data gathered from the composite system.
That is, in addition to the datasets \(\dataset_{1}, \ldots, \dataset_{\numSubmodels}\) used to train the subsystem models, we assume access to a dataset \(\dataset_{\comp} = \{\trajectory_{1}^{\comp}, \ldots, \trajectory_{|\dataset_{\comp}|}^{\comp}\}\) of trajectories \(\trajectory^{\comp} = \{(\state_{\comp, 0}, \controlInput_{\comp, 0}, \timeVar_{0}), \ldots, (\state_{\comp, |\trajectory|}, \controlInput_{\comp, |\trajectory|}, \timeVar_{|\trajectory|})\}\), where \(\state_{\comp, i}\), and \(\controlInput_{\comp, i}\) represent the composite state and control input at time \(\timeVar_{i}\), respectively.

Let \(\compositionMatrix_{\compositionTermParams}(\state_{\comp})\) denote a parametric model of the unknown composition matrix, defined in terms of the parameter vector \(\compositionTermParams\).
Given a collection of pre-trained submodels \(\phnode_{i, \allParams}(\cdot)\) for \(i=1, \ldots, \numSubmodels\), 
the composite model \(\phnode_{\comp, \nnParamsAll_{\comp}}^{\compositionMatrix_{\compositionTermParams}}(\cdot)\) is constructed as defined in \S \ref{sec:composition_known_interconnection}, with the exception of the parametrized composition term \(\compositionMatrix_{\compositionTermParams}(\state_{\comp})\) being used in place of its ground-truth counterpart.

To learn the parameters \(\compositionTermParams\) using the dataset \(\dataset_{\comp}\), we define a loss function \(\loss^{comp}(\compositionTermParams, \nnParamsAll_{\comp}, \dataset_{\comp})\) similarly to as in \eqref{eq:loss_function}, where \(\allParams_{\comp}\) denotes the collection of all of the parameter vectors \(\allParams_{i}\) from the subsystem PHNNs.
In general, \(\compositionMatrix_{\compositionTermParams}(\state_{\comp})\) will be a function of \(\state_{\comp}\) and we parametrize its entries as the output of a neural network.
\(\loss^{comp}(\compositionTermParams, \allParams_{\comp}, \dataset_{\comp})\) can then be minimized by fixing the pre-trained values of \(\allParams_{\comp}\) and performing gradient descent over \(\compositionTermParams\).

However, when the interconnection term \(\structure_{\comp}(\state_{\comp})\) is a constant matrix,  \(\compositionMatrix_{\compositionTermParams}\) may be parameterized as constant skew symmetric matrix.
In such scenarios, so long as the timestep \(\timeVar_{i+1} - \timeVar_{i}\) between the recorded datapoints is sufficiently small, each datapoint yields the following collection of linear equations in the unknown entries of \(\compositionMatrix_{\compositionTermParams}\). 
\begin{equation}
    \frac{\state_{i+1} - \state_{i}}{\timeVar_{i+1} - \timeVar_{i}} - \left[ \Tilde{\structure}_{\comp} - \dissipation_{\comp, \nnParams}(\state_{\comp, i})  \right] \nabla_{\state_{\comp}}\hamiltonian_{\comp, \nnParams}(\state_{\comp, i}) + \inputControlMatrix_{\comp, \nnParams}(\state_{\comp, i}) \controlInput_{\comp, i} \approx  \compositionMatrix_{\compositionTermParams} \nabla_{\state}\hamiltonian_{\comp, \nnParams}(\state_{\comp, i}),
\end{equation}
where we use \(\Tilde{\structure}_{\comp}\) as shorthand for \(\diag(\structure_{1}, \ldots, \structure_{\numSubmodels})\).
Dataset \(\dataset_{\comp}\) thus yields a least squares problem that we solve to obtain the entries of \(\compositionMatrix_{\compositionTermParams}\).

\subsection{Bounding the Errors of Composititions of Port-Hamiltonian Neural Networks}

\label{sec:composing_phnn_error}

\begin{theorem}
    Suppose the true dynamics of each subsystem may be written in port-Hamiltonian form as \(\phsRHS_{i}(\state_i, \controlInput_i)\) for \(i = 1,2,\ldots, \numSubmodels\). Furthermore, suppose the composite system of interest may be represented as a composition \(\phsRHS_{\comp}^{\compositionMatrix}(\state_{\comp}, \controlInput_{\comp})\) of the port-Hamiltonian subsystems defined by the composition term \(\compositionMatrix(\state_{\comp})\).
    Let \(\compositionMatrix_{\compositionTermParams}(\state_{\comp})\) denote the learned composition term and let \(\compositionMatrix_{\compositionTermParams}(\state_{\comp})_{i, j}\) denote the submatrix that defines the interactions between subsystems \(i\) and \(j\).
    Suppose that for every \(i,j = 1,\ldots, k\) with \(i \neq j\), we have 
    \( \norm{\phsRHS_i(\state_i, \controlInput_i) - \phnnRHS_{i, \allParams}(\state_i, \controlInput_i)} \leq \phnnErr_{i}\) 
    and 
    \(\norm{\nabla_{\state_{i}}\hamiltonian_{i}(\state_i) - \nabla_{\state_{i}}\hamiltonian_{i, \nnParams}(\state_i)} \leq \hamiltonianErr_{i}\) 
    for all 
    \(\state_{i} \in \domain_{\state_{i}}\), \(\controlInput_{i} \in \domain_{\controlInput_{i}}\).
    Also suppose that 
    \(\norm{(\compositionMatrix(\state_{\comp})_{i, j} - \compositionMatrix_{\compositionTermParams}(\state_{\comp})_{i, j}) \nabla_{\state_{j}}\hamiltonian_{j}(\state_{j})} \leq \compositionMatrixErr_{i,j}\)
    for all \(\state_{j} \in \domain_{\state_{j}}\) that are consistent with some composite state \(\state_{\comp} \in \domain_{\state_{1}} \times \ldots \times \domain_{\state_{\numSubmodels}}\).
    Then,
    \begin{equation}
        \norm{\phsRHS_{\comp}^{\compositionMatrix}(\state_{\comp}, \controlInput_{\comp}) - \phnnRHS^{\compositionMatrix_{\compositionTermParams}}_{\comp, \allParams_{\comp}}(\state_{\comp}, \controlInput_{\comp})} \leq \sum_{i=1}^{\numSubmodels} \bigl[ \phnnErr_{i} + 2 \sum_{j > i}^{k} (\compositionMatrixErr_{i,j} + \compositionMatrixNormBound_{i,j} \hamiltonianErr_{j})\bigr],
        \label{eq:thm1_err_bound}
    \end{equation}
    for every \(\state_{\comp} \in \domain_{\state_{1}} \times \ldots \times \domain_{\state_{\numSubmodels}}\) and \(\controlInput_{\comp} \in \domain_{\controlInput_{1}} \times \ldots \times \domain_{\controlInput_{\numSubmodels}}\). 
    Here, \(\compositionMatrixNormBound_{i,j} \defeq \max_{\state_{\comp} \in \Omega_{\state_{\comp}}} \norm{\compositionMatrix_{\compositionTermParams}(\state_{\comp})_{i,j}}\) where \(\norm{\compositionMatrix_{\compositionTermParams}(\state_{\comp})_{i,j}} \defeq \sup \{\norm{\compositionMatrix_{\compositionTermParams}(\state_{\comp})_{i,j} y } \textrm{ s.t. } y \in \mathbb{R}^{n_{j}}, \norm{y} = 1\}\) is the matrix norm of \(\compositionMatrix_{\compositionTermParams}(\state_{\comp})_{i, j}\).
    \label{thm:thm_1}
\end{theorem}

In words, Theorem \ref{thm:thm_1} tells us that the prediction error of the composite PHNN is bounded by the error introduced by its component models and the error introduced by the influence that its component models have on each other.
The latter is a function of the error in our estimate of the composition term \(\compositionMatrix_{\compositionTermParams}(\state)\) and in the gradients of the subsystem Hamiltonian functions \(\nabla_{\state_{i}}\hamiltonian_{i, \nnParams}(\state_{i})\).
We note that when \(\compositionMatrix(\state)\) is known a priori this result still holds with all \(\compositionMatrixErr_{i,j} = 0\).
This result ensures that any improvements to the prediction accuracy of the submodels during training will improve the prediction accuracy of the composite model as well.
The proof is provided in Appendix \ref{sec:proofs}.
\section{Numerical Experiments}

For detailed descriptions of the dataset generation, the employed neural network architectures, and the training algorithms used to generate the results, we refer the reader to Appendix \ref{sec:appendix_experimental_details}.
Code to reproduce all experiments is available at \href{https://github.com/cyrusneary/compositional_port_hamiltonian_NNs}{github.com/cyrusneary/compositional\_port\_hamiltonian\_NNs}.

\subsection{Composing Port-Hamiltonian Neural Networks Using a Known Composition Term}

\begin{wrapfigure}{R}{0.5\textwidth}
    \centering
    \input{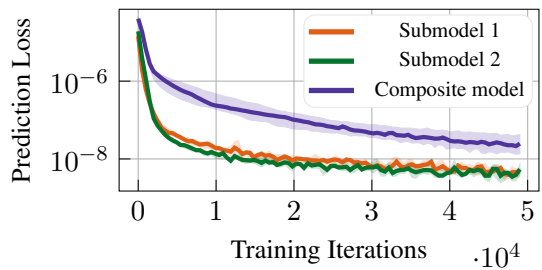}
    \vspace{0mm}
    \caption{
        Prediction loss values throughout training.
        The composite model makes accurate predictions without any training data from the composite system itself.
    }
    \vspace*{-5mm}
    \label{fig:exp_results_known_comp_model_loss}
\end{wrapfigure}

We begin by considering a numerical simulation of the coupled spring-mass-dampers illustrated in Figure \ref{fig:spring_mass_example}.
Recall that this example includes external control forces and nonlinear dissipation.

We first consider the scenario in which the composition term \(\compositionMatrix\) is known a priori.
We learn the subsystem models \(\phnode_{1,\allParams}(\cdot)\) and \(\phnode_{2, \allParams}(\cdot)\) using separate datasets \(\dataset_{1}\) and \(\dataset_{2}\), each of which contains \(100\) trajectories with randomly sampled initial states.
Throughout training we occasionally freeze the network parameters and compose the resulting submodels to obtain a composite model \(\phnode_{\comp, \allParams}^{\compositionMatrix}(\cdot)\).

Figure \ref{fig:exp_results_known_comp_model_loss} illustrates the result of evaluating the prediction error of both submodels, and of the composite model, on separate testing datasets.
The prediction loss measures the average Euclidean distance between the model-predicted future states and the true next states.
We plot the median loss values over \(10\) independent experimental runs, while the shaded regions enclose the \(25^{th}\) and \(75^{th}\) percentiles.
Each testing dataset contains \(20\) system trajectories beginning from randomly sampled initial states, and with different control inputs than the training datasets. 

\paragraph{The composite model accurately predicts dynamics without data from the composite system.}
We emphasize that the composite model's loss values (blue) in Figure \ref{fig:exp_results_known_comp_model_loss} are not the result of training a separate model using composite system data.
Instead, the figure shows the result of learning the subsystem models independently (orange and green), and using a physics-informed composition of the submodels to accurately predict the dynamics of a more complex composite system for which we have no data.
We additionally observe from Figure \ref{fig:exp_results_known_comp_model_loss}, that as the prediction losses of the submodels decrease during training, the loss of the composite model decreases as well.
This observation empirically demonstrates Theorem \ref{thm:thm_1}---
the composite model's error should decrease with the error of the subsystem PHNNs.

Finally, we note that training a baseline model for comparison in this example is not possible; the idea of composing deep learning submodels of dynamical systems is novel, and no monolithic baseline model could possibly learn to predict the composite dynamics without training data.

\subsection{Learning the Interactions Between Subsystem Port-Hamiltonian Neural Networks}

We now proceed to the setting in which the composition term \(\compositionMatrix\) is unknown---we do not have a priori knowledge of how the subsystems are influencing each other.
Instead, as described in \S \ref{sec:composition_unknown_interconnection}, we assume access to some small dataset of \(\dataset_{\comp}\) of observations of the dynamics of the composite system.
Specifically, \(\dataset_{\comp}\) contains only four datapoints, each of which corresponds to a single-timestep transition beginning from a randomly sampled composite state.
Using these datapoints, along with the pretrained submodels \(\phnode_{1, \allParams}(\cdot)\) and \(\phnode_{2,\allParams}(\cdot)\), we learn \(\compositionMatrix_{\compositionTermParams}\) by solving the least squares problem described in \S \ref{sec:composition_unknown_interconnection}.

\begin{figure}
    \centering
    \input{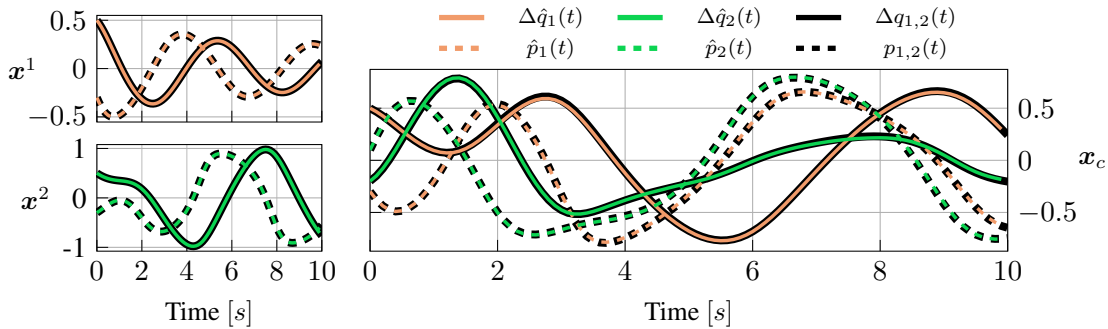}
    \caption{
        The composite PHNN accurately predicts state trajectories, even when the unknown composition term \(\compositionMatrix_{\compositionTermParams}\) is inferred using only four transition datapoints.
        Left: Predicted subsystem dynamics.
        Right: Predicted composite system dynamics.
        }
    \label{fig:exp_results_composite_phnode_predicted_trajectory}
\end{figure}

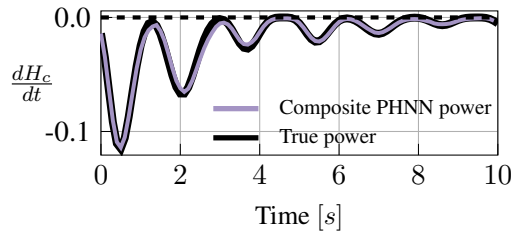
\begin{wrapfigure}{R}{0.5\textwidth}
\begin{tikzpicture}

\definecolor{darkgray176}{RGB}{176,176,176}
\definecolor{gray141124134}{RGB}{141,124,134}
\definecolor{lightgray204}{RGB}{204,204,204}

\setlength{\thinLines}{1.5pt}
\setlength{\thickLines}{3.0pt}

\begin{axis}[
legend cell align={left},
legend style={
  fill opacity=0.8,
  font=\footnotesize,
  draw opacity=1,
  text opacity=1,
  at={(0.97,0.03)},
  anchor=south east,
  draw=lightgray204
},
tick align=inside,
tick pos=left,
x grid style={darkgray176},
xlabel={\small Time \([s]\)},
xmajorgrids,
xmin=-0.0, xmax=10.0,
xtick style={color=black},
y grid style={darkgray176},
ylabel={\(\frac{dH_{\comp}}{dt}\)},
ymajorgrids,
ytick={-0.1, 0.0},
yticklabels={-0.1, 0.0},
ylabel style={rotate=-90},
ymin=-0.12070056486925, ymax=0.00574692552100657,
ytick style={color=black},
ylabel shift=-0.4cm,
height=3.5cm,
width=0.45*\textwidth,
]

\addplot [line width=\thickLines, black]
table {%
0 -0.0139200016856194
0.0999999940395355 -0.0350818671286106
0.199999988079071 -0.0637532845139503
0.299999982118607 -0.0917849540710449
0.399999976158142 -0.110454611480236
0.5 -0.114952951669693
0.599999964237213 -0.105743050575256
0.699999988079071 -0.0870493203401566
0.799999952316284 -0.0643760561943054
0.899999976158142 -0.042535237967968
1 -0.024686872959137
1.10000002384186 -0.0122718662023544
1.19999992847443 -0.00549342064186931
1.29999995231628 -0.00396352633833885
1.39999997615814 -0.00717362435534596
1.5 -0.0145866526290774
1.59999990463257 -0.0253870654851198
1.69999992847443 -0.0381604060530663
1.79999995231628 -0.0508311465382576
1.89999997615814 -0.0610155500471592
2 -0.0666669458150864
2.09999990463257 -0.0667064934968948
2.20000004768372 -0.061338946223259
2.29999995231628 -0.0519208982586861
2.39999985694885 -0.0404621511697769
2.5 -0.0289869531989098
2.59999990463257 -0.0190061945468187
2.70000004768372 -0.0112775461748242
2.79999995231628 -0.00590335065498948
2.89999985694885 -0.00267573515884578
3 -0.00145746837370098
3.09999990463257 -0.00234774523414671
3.19999980926514 -0.0054937326349318
3.29999995231628 -0.0106651186943054
3.39999985694885 -0.016934160143137
3.5 -0.0227739121764898
3.59999990463257 -0.0265791211277246
3.69999980926514 -0.0272995326668024
3.79999995231628 -0.0248221233487129
3.89999985694885 -0.0199403427541256
4 -0.0139911659061909
4.09999990463257 -0.00836068391799927
4.19999980926514 -0.00405243411660194
4.29999971389771 -0.00144655257463455
4.40000009536743 -0.000306521833408624
4.5 -2.20244510273915e-05
4.59999990463257 -6.87678550548299e-07
4.69999980926514 -4.24967984145042e-05
4.79999971389771 -0.000489918049424887
4.90000009536743 -0.00201778719201684
5 -0.00513352360576391
5.09999990463257 -0.00969037972390652
5.19999980926514 -0.0147622190415859
5.29999971389771 -0.0189936459064484
5.40000009536743 -0.0212029479444027
5.5 -0.0208675358444452
5.59999990463257 -0.018254354596138
5.69999980926514 -0.0142070548608899
5.79999971389771 -0.00976789370179176
5.90000009536743 -0.00583044020459056
6 -0.00294404313899577
6.09999990463257 -0.00129462184850127
6.19999980926514 -0.000813550315797329
6.29999971389771 -0.00132860615849495
6.39999961853027 -0.00267168181017041
6.5 -0.00469516357406974
6.59999990463257 -0.00720777688547969
6.69999980926514 -0.00989504344761372
6.79999971389771 -0.0123038729652762
6.89999961853027 -0.0139314625412226
7 -0.0143880965188146
7.09999990463257 -0.0135489869862795
7.19999980926514 -0.0116112427785993
7.29999971389771 -0.00902351085096598
7.39999961853027 -0.00632353406399488
7.5 -0.0039606555365026
7.59999990463257 -0.00218103430233896
7.69999980926514 -0.00101792556233704
7.79999971389771 -0.000377015414414927
7.89999961853027 -0.000159973598783836
8 -0.000348597532138228
8.09999942779541 -0.000994782545603812
8.19999980926514 -0.0021220687776804
8.30000019073486 -0.00361001421697438
8.39999961853027 -0.00515284016728401
8.5 -0.00633774464949965
8.59999942779541 -0.00680726720020175
8.69999980926514 -0.00641420343890786
8.80000019073486 -0.00528464047238231
8.89999961853027 -0.00376064726151526
9 -0.0022561764344573
9.09999942779541 -0.00109590170904994
9.19999980926514 -0.000408179970690981
9.30000019073486 -0.000116658964543603
9.39999961853027 -3.34848227794282e-05
9.5 -9.92512013908708e-06
9.59999942779541 -6.48190834908746e-05
9.69999980926514 -0.000408734165830538
9.80000019073486 -0.00132882408797741
9.89999961853027 -0.00298517686314881
10 -0.00524078961461782
};
\addplot [line width=\thinLines, lightCompositeModelColor]
table {%
0 -0.013973169028759
0.1 -0.0346340425312519
0.2 -0.0628266260027885
0.3 -0.090359091758728
0.4 -0.109148614108562
0.5 -0.114037923514843
0.6 -0.105081714689732
0.7 -0.0868151485919952
0.8 -0.0654834136366844
0.900000000000001 -0.0445650219917297
1 -0.0284041855484247
1.1 -0.0174588356167078
1.2 -0.0111896730959415
1.3 -0.00795926060527563
1.4 -0.00662732031196356
1.5 -0.0118448464199901
1.6 -0.0201999861747026
1.7 -0.0319472849369049
1.8 -0.0444855391979218
1.9 -0.0555269755423069
2 -0.0621195323765278
2.1 -0.0652624219655991
2.2 -0.0624694749712944
2.29999999999999 -0.0555333495140076
2.39999999999999 -0.0461650341749191
2.49999999999999 -0.0361791290342808
2.59999999999999 -0.0269878283143044
2.69999999999999 -0.0193584226071835
2.79999999999998 -0.0133377937600017
2.89999999999998 -0.00822651665657759
2.99999999999998 -0.00522079551592469
3.09999999999998 -0.00455807149410248
3.19999999999998 -0.00605915300548077
3.29999999999997 -0.00989233329892159
3.39999999999997 -0.0151702677831054
3.49999999999997 -0.0203116647899151
3.59999999999997 -0.023325240239501
3.69999999999997 -0.0238405559211969
3.79999999999996 -0.021718867123127
3.89999999999996 -0.0181902572512627
3.99999999999996 -0.0128988083451986
4.09999999999996 -0.00784092210233212
4.19999999999995 -0.00421415409073234
4.29999999999995 -0.00213774223811924
4.39999999999995 -0.00105392176192254
4.49999999999995 -0.000849661359097809
4.59999999999995 -0.000835623068269342
4.69999999999994 -0.000778334739152342
4.79999999999994 -0.00111992890015244
4.89999999999994 -0.00197117612697184
4.99999999999994 -0.00479769334197044
5.09999999999994 -0.00873356685042381
5.19999999999993 -0.0134673267602921
5.29999999999993 -0.0178331881761551
5.39999999999993 -0.0206225756555796
5.49999999999993 -0.0211074184626341
5.59999999999993 -0.0192925482988358
5.69999999999992 -0.0158237628638744
5.79999999999992 -0.0116462036967278
5.89999999999992 -0.007333695422858
5.99999999999992 -0.00398978218436241
6.09999999999991 -0.0019464276265353
6.19999999999991 -0.00129316013772041
6.29999999999991 -0.00162895163521171
6.39999999999991 -0.0027939323335886
6.49999999999991 -0.00464751524850726
6.5999999999999 -0.00699487281963229
6.6999999999999 -0.00937642063945532
6.7999999999999 -0.0114259161055088
6.8999999999999 -0.0132285142317414
6.9999999999999 -0.0136859575286508
7.09999999999989 -0.0128982337191701
7.19999999999989 -0.0111641818657517
7.29999999999989 -0.00907442066818476
7.39999999999989 -0.0066345245577395
7.49999999999988 -0.00445496058091521
7.59999999999988 -0.0029436550103128
7.69999999999988 -0.00193547457456589
7.79999999999988 -0.00155623420141637
7.89999999999988 -0.00140790711157024
7.99999999999987 -0.00156442867591977
8.09999999999987 -0.00207898626103997
8.19999999999987 -0.00298613915219903
8.29999999999987 -0.00381792918778956
8.39999999999987 -0.00523333391174674
8.49999999999986 -0.00585221219807863
8.59999999999986 -0.00593563029542565
8.69999999999986 -0.00576545903459191
8.79999999999986 -0.00525894062593579
8.89999999999985 -0.00420592725276947
8.99999999999985 -0.00298127136193216
9.09999999999985 -0.0020343498326838
9.19999999999985 -0.00144198490306735
9.29999999999985 -0.00112113170325756
9.39999999999984 -0.000922883802559227
9.49999999999984 -0.00076344091212377
9.59999999999984 -0.000608281581662595
9.69999999999984 -0.00075407128315419
9.79999999999984 -0.00143317854963243
9.89999999999983 -0.00286035193130374
};

\addplot [ultra thick, black, dashed]
table {%
0.0 0.0
10.0 0.0
};

\end{axis}

\begin{customlegend}[
    legend cell align={left},
    legend columns=1, 
    legend style={
        align=left, 
        column sep=1ex, 
        font=\scriptsize, 
        draw=none,
        fill=none,
        inner xsep=0.3mm,
        inner ysep=0.3mm,
        rounded corners=1.5mm,
        at={(53.0mm, 8.5mm)},
    }, 
    legend entries={
        Composite PHNN power\\ 
        True power\\ 
    }
]
\addlegendimage{line width=2pt, lightCompositeModelColor}
\addlegendimage{line width=2pt, black}
\end{customlegend}

\end{tikzpicture}
    \vspace{-3mm}
    \caption{Time derivative of the system's Hamiltonian with no external control inputs.}
    \vspace*{-0.5cm}
    \label{fig:exp_results_composite_phnode_predicted_power}
\end{wrapfigure}

\paragraph{We accurately learn the unknown composition term using a negligible amount of data.}
Figure \ref{fig:exp_results_composite_phnode_predicted_trajectory} illustrates state trajectories predicted by the subsystem, and composite, PHNNs.
In all of the subplots, the true dynamics (black) are matched very closely by the PHNN-predicted dynamics (orange and green);
we accurately learn the composition term \(\compositionMatrix_{\compositionTermParams}\) using only four datapoints from the composite system.
Monolithic approaches cannot learn effective models from such a limited dataset.

\paragraph{The learned composite model enjoys properties of port-Hamiltonian systems.}
Figure \ref{fig:exp_results_composite_phnode_predicted_power} illustrates the time derivative of the learned composite Hamiltonian along a predicted trajectory without external control inputs.
We note that for the entire trajectory, \(d \hamiltonian_{\comp, \nnParams} / dt \leq 0\),
which provides an empirical demonstration of the cyclo-passivity of the composite PHNN discussed in \S \ref{sec:phnns}.

\subsection{Predicting the Dynamics of Ten Interacting Subsystems Without Additional Training}

\begin{wrapfigure}{R}{0.5\textwidth}
    \vspace{-5mm}
    \input{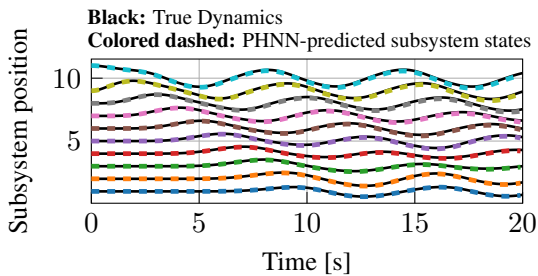}
    \vspace{-3mm}
    \caption{The composite PHNN accurately predicts the dynamics of \(10\) interacting subsystems with no additional training.}
    \vspace*{-5mm}
    \label{fig:exp_10_subsystems}
\end{wrapfigure}

We now consider ten interacting spring-mass-damper systems, most of which are identical to subsystem \(1\) from Figure \ref{fig:spring_mass_example}, while the rest match subsystem \(2\).
A sinusoidal force acts on one end of the chain.
Figure \ref{fig:exp_10_subsystems} illustrates the results of composing the previously trained PHNN submodels to predict this system's dynamics.
The predicted state trajectory of each subsystem is plotted as a separate colored curve.

The composite PHNN accurately predicts the wave-like propagation of energy between the subsystems (the colored PHNN predictions overlay the true dynamics in black).
We emphasize that this behavior is captured without any additional training of the subsystem models, and without access to any data from the ten-component system; PHNNs can be composed in a modular fashion to simulate entirely new complex systems.
\section{Conclusions}

In this work we present a framework, algorithms, and theoretical results for the compositional learning of dynamical system models via port-Hamiltonian neural networks (PHNNs).
This work presents a first step towards learning modular neural network parametrizations of control systems that can be trained and tested independently, and that can be re-used in new contexts.
We demonstrate that by using the structure of port-Hamiltonian systems as inductive bias, we may independently learn submodels on data generated by relatively simple subsystems, and then accurately predict the dynamics of more complex composite systems while using little to no data from the composite system itself.
Future work will aim to learn compositional PHNNs from video observations, and also to use the compositional models for control.

\acks{This work was supported in part by AFOSR FA9550-19-1-0005, ARO W911NF-20-1-0140, and NSF 2214939.}

\bibliography{bibliography.bib}

\newpage

\setcounter{theorem}{0}

\begin{center} 
    \begin{Large}
        \textbf{Compositional Learning of Dynamical System Models Using Port-Hamiltonian Neural Networks: Supplementary Material}
    \end{Large}
\end{center}

\appendix
\section{Proof of Theorem 1}
\label{sec:proofs}

\begin{theorem}
    Suppose the true dynamics of each subsystem may be written in port-Hamiltonian form as \(\phsRHS_{i}(\state_i, \controlInput_i)\) for \(i = 1,2,\ldots, \numSubmodels\). Furthermore, suppose the composite system of interest may be represented as a composition \(\phsRHS_{\comp}^{\compositionMatrix}(\state_{\comp}, \controlInput_{\comp})\) of the port-Hamiltonian subsystems defined by the composition term \(\compositionMatrix(\state_{\comp})\).
    Let \(\compositionMatrix_{\compositionTermParams}(\state_{\comp})\) denote the learned composition term and let \(\compositionMatrix_{\compositionTermParams}(\state_{\comp})_{i, j}\) denote the submatrix that defines the interactions between subsystems \(i\) and \(j\).
    Suppose that for every \(i,j = 1,\ldots, k\) with \(i \neq j\), we have 
    \( \norm{\phsRHS_i(\state_i, \controlInput_i) - \phnnRHS_{i, \allParams}(\state_i, \controlInput_i)} \leq \phnnErr_{i}\) 
    and 
    \(\norm{\nabla_{\state_{i}}\hamiltonian_{i}(\state_i) - \nabla_{\state_{i}}\hamiltonian_{i, \nnParams}(\state_i)} \leq \hamiltonianErr_{i}\) 
    for all 
    \(\state_{i} \in \domain_{\state_{i}}\), \(\controlInput_{i} \in \domain_{\controlInput_{i}}\).
    Also suppose that 
    \(\norm{(\compositionMatrix(\state_{\comp})_{i, j} - \compositionMatrix_{\compositionTermParams}(\state_{\comp})_{i, j}) \nabla_{\state_{j}}\hamiltonian_{j}(\state_{j})} \leq \compositionMatrixErr_{i,j}\)
    for all \(\state_{j} \in \domain_{\state_{j}}\) that are consistent with some composite state \(\state_{\comp} \in \domain_{\state_{1}} \times \ldots \times \domain_{\state_{\numSubmodels}}\).
    Then,
    \begin{equation}
        \norm{\phsRHS_{\comp}^{\compositionMatrix}(\state_{\comp}, \controlInput_{\comp}) - \phnnRHS^{\compositionMatrix_{\compositionTermParams}}_{\comp, \allParams_{\comp}}(\state_{\comp}, \controlInput_{\comp})} \leq \sum_{i=1}^{\numSubmodels} \bigl[ \phnnErr_{i} + 2 \sum_{j > i}^{k} (\compositionMatrixErr_{i,j} + \compositionMatrixNormBound_{i,j} \hamiltonianErr_{j})\bigr],
        \label{eq:thm1_err_bound}
    \end{equation}
    for every \(\state_{\comp} \in \domain_{\state_{1}} \times \ldots \times \domain_{\state_{\numSubmodels}}\) and \(\controlInput_{\comp} \in \domain_{\controlInput_{1}} \times \ldots \times \domain_{\controlInput_{\numSubmodels}}\). 
    Here, \(\compositionMatrixNormBound_{i,j} \defeq \max_{\state_{\comp} \in \Omega_{\state_{\comp}}} \norm{\compositionMatrix_{\compositionTermParams}(\state_{\comp})_{i,j}}\) where \(\norm{\compositionMatrix_{\compositionTermParams}(\state_{\comp})_{i,j}} \defeq \sup \{\norm{\compositionMatrix_{\compositionTermParams}(\state_{\comp})_{i,j} y } \textrm{ s.t. } y \in \mathbb{R}^{n_{j}}, \norm{y} = 1\}\) is the matrix norm of \(\compositionMatrix_{\compositionTermParams}(\state_{\comp})_{i, j}\).
    \label{thm:thm_1}
\end{theorem}

\begin{proof}
We begin by expanding the expression for the error in the composite PHNN.
\begin{align}
    ||\phsRHS_{\comp}(\state_{\comp},& \controlInput_{\comp}) - \phnnRHS^{\compositionMatrix_{\compositionTermParams}}_{\comp, \allParams_{\comp}}(\state_{\comp}, \controlInput_{\comp})|| \\
    & = ||\left[ \Tilde{\structure}_{\comp}(\state_{\comp}) + \compositionMatrix(\state_{\comp}) - \dissipation_{\comp}(\state_{\comp}) \right] \nabla_{\state_{\comp}} \hamiltonian_{\comp}(\state_{\comp}) + \inputControlMatrix_{\comp}(\state_{\comp}) \controlInput_{\comp} \label{eq:expanded_phnn_rhs_err} \\ 
    & \;\;\;\;\; - \left[\Tilde{\structure}_{\comp}(\state_{\comp}) + \compositionMatrix_{\compositionTermParams}(\state_{\comp}) + \dissipation_{\comp, \nnParams}(\state_{\comp})\right]\nabla_{\state_{\comp}}\hamiltonian_{\comp, \nnParams}(\state_{\comp}) + \inputControlMatrix_{\comp, \nnParams}(\state_{\comp}) \controlInput_{\comp}|| \nonumber \\
    & \leq ||\left[ \Tilde{\structure}_{\comp}(\state_{\comp}) - \dissipation_{\comp}(\state_{\comp}) \right] \nabla_{\state_{\comp}} \hamiltonian_{\comp}(\state_{\comp}) + \inputControlMatrix_{\comp}(\state_{\comp}) \controlInput_{\comp} \label{eq:decomposed_phnn_rhs_err}\\ 
    & \;\;\;\;\; - \left[\Tilde{\structure}_{\comp}(\state_{\comp}) - \dissipation_{\comp, \nnParams}(\state_{\comp})\right]\nabla_{\state_{\comp}}\hamiltonian_{\comp, \nnParams}(\state_{\comp}) - \inputControlMatrix_{\comp, \nnParams}(\state_{\comp}) \controlInput_{\comp}|| \nonumber \\
    & \;\;\; + \norm{\compositionMatrix(\state_{\comp}) \nabla_{\state_{\comp}} \hamiltonian_{\comp}(\state_{\comp}) - \compositionMatrix_{\compositionTermParams}(\state_{\comp}) \nabla_{\state_{\comp}} \hamiltonian_{\comp, \nnParams}(\state_{\comp})} \nonumber \\
    & \leq \sum_{i = 1}^{\numSubmodels} \bigl\{ || \left[\structure_{i}(\state_i) - \dissipation_{i}(\state_{i})\right] (\nabla_{\state_{\comp}} \hamiltonian_{\comp}(\state_{\comp}))_{\indexSet_{i}} + \inputControlMatrix_{i}(\state_{i})\controlInput_{i} \bigr. \label{eq:expanded_composite_err} \\
    & \bigl. \;\;\;\;\;\;\;\;\; - \left[ \structure_{i, \nnParams}(\state_{i}) - \dissipation_{i,\nnParams}(\state_{i}) \right] (\nabla_{\state_{\comp}} \hamiltonian_{\comp,\nnParams}(\state_{\comp}))_{\indexSet_{i}} + \inputControlMatrix_{i, \nnParams}(\state_{i}) \controlInput_{i} || \bigr\} \nonumber \\
    & \;\;\; + \norm{\compositionMatrix(\state_{\comp}) \nabla_{\state_{\comp}} \hamiltonian_{\comp}(\state_{\comp}) - \compositionMatrix_{\compositionTermParams}(\state_{\comp}) \nabla_{\state_{\comp}} \hamiltonian_{\comp, \nnParams}(\state_{\comp})} \nonumber
\end{align}

Recall that we define \(\Tilde{\structure}_{\comp}(\state_{\comp}) \defeq \diag(\structure_{1}(\state_{1}), \ldots, \structure_{\numSubmodels}(\state_{\numSubmodels}))\).
\Eqref{eq:decomposed_phnn_rhs_err} follows from \eqref{eq:expanded_phnn_rhs_err} by the triangle inequality.
\Eqref{eq:expanded_composite_err} follows from \eqref{eq:decomposed_phnn_rhs_err} again by the triangle inequality and due to the block-diagonal structure of \(\Tilde{\structure}_{\comp}(\state_{\comp})\), \(\dissipation_{\comp}(\state_{\comp})\), and \(\inputControlMatrix_{\comp}(\state_{\comp})\).

In \eqref{eq:expanded_composite_err} we use \(\indexSet_{i}\) to denote the set of indexes corresponding to subsystem \(i\) within the composite vector \(\state_{\comp}\).
More precisely, \(\indexSet_{i} \defeq \{j | j > \stateDim_{1} + \ldots + \stateDim_{i-1} \textrm{ and } j \leq \stateDim_{1} + \ldots + \stateDim_{i}\}\).
We thus use \((\nabla_{\state_{\comp}} \hamiltonian_{\comp}(\state_{\comp}))_{\indexSet_{i}}\) to denote the vector in \(\mathbb{R}^{\stateDim_{i}}\) that results from taking the elements from \(\nabla_{\state_{\comp}} \hamiltonian_{\comp}(\state_{\comp})\) indexed by \(\indexSet_{i}\) and discarding the rest of the vector. 
Then, by definition of \(\hamiltonian_{\state_{\comp}}(\state_{\comp})\), \(\nabla_{\state_{\comp}} \hamiltonian_{\comp}(\state_{\comp}) = \sum_{j=1}^{\numSubmodels} \nabla_{\state_{\comp}} \hamiltonian_{j}(\state_{j}) \).
For every \(i \neq j\), we have that \((\nabla_{\state_{\comp}} \hamiltonian_{j}(\state_{j}))_{\indexSet_{i}} = \vzero \in \mathbb{R}^{\stateDim_{i}}\) because of our assumption that the subsystem Hamiltonian \(\hamiltonian_{j}(\cdot)\) only depends on \(\state_{j}\). 
Furthermore, using the same reasoning it must be true that \((\nabla_{\state_{\comp}} \hamiltonian_{j}(\state_{j}))_{\indexSet_{j}} = \nabla_{\state_{j}} \hamiltonian_{j}(\state_{j})\).
So, we conclude that 
\begin{equation}
    \label{eq:hamiltonian_grad_index_relationship}
    (\nabla_{\state_{\comp}} \hamiltonian_{\comp}(\state_{\comp}) )_{\indexSet_{i}} = \sum_{j=1}^{\numSubmodels} (\nabla_{\state_{\comp}} \hamiltonian_{j}(\state_{j}))_{\indexSet_{i}} = \nabla_{\state_{i}}\hamiltonian_{i}(\state_{i}).
\end{equation}
By combining \eqref{eq:hamiltonian_grad_index_relationship} with \eqref{eq:expanded_composite_err}, we have
\begin{align}
    ||\phsRHS_{\comp}(\state_{\comp}, \controlInput_{\comp}) - \phnnRHS^{\compositionMatrix_{\compositionTermParams}}_{\comp, \allParams_{\comp}}(\state_{\comp}, \controlInput_{\comp})|| & \leq \sum_{i=1}^{\numSubmodels} ||\phsRHS_{i}(\state_i, \controlInput_{i}) - \phsRHS_{i, \allParams}(\state_{i}, \controlInput_{i}) || \\ 
    & \;\;\;\;\; + \norm{\compositionMatrix(\state_{\comp}) \nabla_{\state_{\comp}} \hamiltonian_{\comp}(\state_{\comp}) - \compositionMatrix_{\compositionTermParams}(\state_{\comp}) \nabla_{\state_{\comp}} \hamiltonian_{\comp, \nnParams}(\state_{\comp})} \nonumber \\
    & \leq \sum_{i=1}^{\numSubmodels} \phnnErr_{i} + \norm{\compositionMatrix(\state_{\comp}) \nabla_{\state_{\comp}} \hamiltonian_{\comp}(\state_{\comp}) - \compositionMatrix_{\compositionTermParams}(\state_{\comp}) \nabla_{\state_{\comp}} \hamiltonian_{\comp, \nnParams}(\state_{\comp})}. \label{eq:composite_err_triangle_ineq}
\end{align}
We bound the final term in the right-hand side of \eqref{eq:composite_err_triangle_ineq} as follows.
\begin{align}
    || \compositionMatrix(\state_{\comp}) & \nabla_{\state_{\comp}} \hamiltonian_{\comp}(\state_{\comp}) - \compositionMatrix_{\compositionTermParams}(\state_{\comp}) \nabla_{\state_{\comp}} \hamiltonian_{\comp, \nnParams}(\state_{\comp}) || \label{eq:initial_composition_term} \\
    & \leq ||(\compositionMatrix(\state_{\comp}) - \compositionMatrix_{\compositionTermParams}(\state_{\comp})) \nabla_{\state_{\comp}} \hamiltonian_{\comp}(\state_{\comp})|| + || \compositionMatrix_{\compositionTermParams}(\state_{\comp}) (\nabla_{\state_{\comp}} \hamiltonian_{\comp}(\state_{\comp}) - \nabla_{\state_{\comp}} \hamiltonian_{\comp, \nnParams}(\state_{\comp}))|| \label{eq:triangle_inequality_composition_term} \\
    & \leq \sum_{\substack{i=1 \\ j\neq i}}^{\numSubmodels} \bigl\{ ||(\compositionMatrix(\state_{\comp})_{\indexSet_{i}, \indexSet_{j}} - \compositionMatrix_{\compositionTermParams}(\state_{\comp})_{\indexSet_{i}, \indexSet_{j}}) \nabla_{\state_{j}} \hamiltonian_{j}(\state_{j})|| \bigr. \label{eq:bound_composite_blocks_term} \\
    & \;\;\;\;\;\;\;\;\;\; \bigl. + ||\compositionMatrix_{\compositionTermParams}(\state_{\comp})_{\indexSet_{i}, \indexSet_{j}}(\nabla_{\state_{j}}\hamiltonian_{j}(\state_{j}) - \nabla_{\state_{j}}\hamiltonian_{j, \nnParams}(\state_{j}))|| \bigr\} \\
    &\leq 2  \sum_{\substack{i=1 \\ j > i}}^{k} \bigl\{ \compositionMatrixErr_{i,j} + ||\compositionMatrix_{\compositionTermParams}(\state_{\comp})|| \; ||\nabla_{\state_{j}}\hamiltonian_{j}(\state_{j}) - \nabla_{\state_{j}}\hamiltonian_{j, \nnParams}(\state_{j})|| \bigr\} \\ 
    & \leq 2 \sum_{\substack{i=1 \\ j > i}}^{k} \compositionMatrixErr_{i,j} + \compositionMatrixNormBound_{i,j} \hamiltonianErr_{j} \label{eq:composite_term_blocks_final}
\end{align}

\Eqref{eq:triangle_inequality_composition_term} follows from \eqref{eq:initial_composition_term} by adding and subtracting \(\compositionMatrix_{\compositionTermParams}(\state_{\comp})\nabla_{\state_{\comp}}\hamiltonian_{\comp}(\state_{\comp})\), regrouping terms, and applying the triangle inequality.
We obtain \eqref{eq:bound_composite_blocks_term} by rewriting the matrix multiplications from \eqref{eq:triangle_inequality_composition_term} using the submatrices \(\compositionMatrix(\state_{\comp})_{\indexSet_{i}, \indexSet_{j}}\) of \(\compositionMatrix(\state_{\comp})\) (which correspond to the interactions between subsystem \(i\) and \(j\)).
We exclude \(\compositionMatrix(\state_{\comp})_{\indexSet_{i}, \indexSet_{i}}\) from the sum in \eqref{eq:bound_composite_blocks_term} because, by definition, \(\compositionMatrix(\state_{\comp})_{\indexSet_{i}, \indexSet_{i}} = \vzero\) for all \(i=1,2,\ldots,\numSubmodels\).

Finally, by combining \eqref{eq:composite_term_blocks_final} with \eqref{eq:composite_err_triangle_ineq}, we obtain the desired result.

\end{proof}
\newpage
\section{Additional Experimental Details}
\label{sec:appendix_experimental_details}

\subsection{Simulation Details}

As described in \S \ref{sec:illustrative_example}, the dynamics of the individual spring-mass-damper systems are given by \eqref{eq:state_input_output_ph_form_state} with \(\hamiltonian_{i}(\state_{i}) = \frac{\generalizedMomentum_{i}^{2}}{2 \mass_{i}} + \frac{\springConstant \Delta \generalizedPosition_{i}^{2}}{2}\) and with

\begin{equation}
    \structure_i = \begin{bmatrix}0 & 1 \\ -1 & 0\end{bmatrix}, \;\;
    \dissipation_i(\state_i) = \begin{bmatrix}0 & 0 \\ 0 & \dampingConstant_{i} \frac{\generalizedMomentum_{i}^2}{\mass_{i}^{2}} \end{bmatrix}, \;\;
    \inputControlMatrix_{i} = \begin{bmatrix} 0 \\ 1 \end{bmatrix},
\end{equation}

for each subsystem.
Here, the system states are defined as \(\state_{i} = (\Delta \generalizedPosition_{i}, \generalizedMomentum_{i})\) for \(i=1,2\), where \(\Delta \generalizedPosition_{i}\) is the elongation of the spring and \(\generalizedMomentum_{i}\) is the momentum of the mass.

The dynamics of the coupled system are then given by \eqref{eq:state_input_output_ph_form_state} where the Hamiltonian \(\hamiltonian_{\comp}(\state_{\comp})\) of the composite system is given by \(\hamiltonian_{\comp}(\state_{\comp}) = \hamiltonian_{1}(\state_1) + \hamiltonian_{2}(\state_{2})\) and the composite dynamics may be written in the form of \eqref{eq:state_input_output_ph_form_state} with
\begin{equation}
    \structure_{\comp} = \begin{bmatrix}0 & 1 & 0 & 0 \\ -1 & 0 & 1 & 0 \\ 0 & -1 & 0 & 1 \\ 0 & 0 & -1 & 0\end{bmatrix}, \;\;
    \dissipation_{\comp}(\state_{\comp}) = \begin{bmatrix} 0 & 0 & 0  & 0 \\ 0 & \dampingConstant_{1} \frac{\generalizedMomentum_{1}^2}{\mass_{1}^{2}}  & 0 & 0 \\ 0 & 0 & 0 & 0\\ 0 & 0 & 0  & \dampingConstant_{2} \frac{\generalizedMomentum_{2}^2}{\mass_{2}^{2}} \end{bmatrix}, \;\;
    \inputControlMatrix_{\comp} = \begin{bmatrix} 0 & 0 \\ 1 & 0 \\ 0 & 0 \\ 0 & 1 \end{bmatrix}.
    \label{eq:composite_spring_mass_equations}
\end{equation}

In our numerical experiments we use a fixed timestep RK4 integration scheme with a timestep of \(\dt = 0.01\) for all simulations.
We define subsystem \(1\) to have parameters \(m_{1} = 1.0\), \(k_{1} = 1.2\), and \(b_{1} = 1.7\).
Meanwhile, we define subsystem \(2\) to have parameters \(m_{2} = 1.0\), \(k_{2} = 1.5\), and \(b_{1.7}\).
The training datasets \(\dataset_{1}\) and \(\dataset_{2}\) correspond to 100 trajectories, of 500 timesteps each, generated by simulating the dynamics from random initial states uniformly sampled from \((\Delta \generalizedPosition_{i}, \generalizedMomentum_{i}) \in [-1.0, 1.0]\). 
When generating dataset \(\dataset_{2}\), we additionally include a control input force that varies sinusoidally in time.
We similarly generate testing datasets by simulating 20 trajectories from each subsystem beginning from initial states randomly sampled from the same distribution.
To generate the testing dataset for the composite system \(\dataset_{\comp}\), we simulate 20 trajectories of the coupled dynamics from randomly sampled initial states \((\Delta \generalizedPosition_{1}, \generalizedMomentum_{1}, \Delta \generalizedPosition_{2}, \generalizedMomentum_{2}) \in [-1.0, 1.0, -1.0, 1.0]\) while applying a sinusoidal forcing function to subsystem \(2\).

\subsection{Neural Network Implementations}

All numerical experiments were implemented using the python library \textit{Jax} \citep{jax2018github}, in order to take advantage of its automatic differentiation and just-in-time compilation features.
All experiments were run locally on a desktop computer with a \(12^{th}\) generation Intel i9 CPU, an Nvidia RTX A4000 GPU, and with 32 GB of RAM.

The Hamiltonian \(\hamiltonian_{\nnParams}(\state)\) and dissipation \(\dissipation_{\nnParams}(\state)\) neural networks are implemented as multilayer perceptrons with TANH activation functions and with two hidden layers of \(32\) units each.
Meanwhile, we parametrize the entries of the constant control input matrix \(\inputControlMatrix_{\nnParams}(\state)\) directly.

We train all models using ADAM \citep{kingma2014adam} for a fixed number of training steps with a learning rate of \(\alpha = 10^{-3}\). 
We use a minibatch size of \(32\). 
We do not include any additional regularization terms in the loss function.

\end{document}